\documentclass{article}
\usepackage{spconf,amsmath,graphicx}
\usepackage{cite}
\usepackage{amsmath,amssymb,amsfonts}
\usepackage{algorithmic}
\usepackage{textcomp}
\usepackage{xcolor}
\usepackage{enumitem}

\title{Scalable AI Generative Content for Vehicular Network Semantic Communication}
\twoauthors
 {Hao Feng, Yi Yang}
	{Tsinghua University\\
	Beijing, China}
 {Zhu Han}
	{University of Houston\\
 Houston, TX 77004}
\begin{document}
%
\maketitle
\begin{abstract}
Perceiving vehicles in a driver's blind spot is vital for safe driving. The detection of potentially dangerous vehicles in these blind spots can benefit from vehicular network semantic communication technology. However, efficient semantic communication involves a trade-off between accuracy and delay, especially in bandwidth-limited situations. This paper unveils a scalable Artificial Intelligence Generated Content (AIGC) system that leverages an encoder-decoder architecture. This system converts images into textual representations and reconstructs them into quality-acceptable images, optimizing transmission for vehicular network semantic communication. Moreover, when bandwidth allows, auxiliary information is integrated. The encoder-decoder aims to maintain semantic equivalence with the original images across various tasks. Then the proposed approach employs reinforcement learning to enhance the reliability of the generated contents. Experimental results suggest that the proposed method surpasses the baseline in perceiving vehicles in blind spots and effectively compresses communication data. While this method is specifically designed for driving scenarios, this encoder-decoder architecture also holds potential for wide use across various semantic communication scenarios.
\end{abstract}


\section{Introduction}
\label{sec:intro}

Artificial Intelligence Generated Content (AIGC) \cite{goodfellow2014generative, ramesh2021zero, Unleashing_minrui, 10177684} leverages advanced machine learning and deep learning techniques, enabling computers to produce vast amounts of textual, graphical, auditory, and video content from minimal information. However, challenges persist in AIGC applications, including the generation of task-specific, contextually relevant content and assessment of output quality.

One practical application of AIGC is in vehicular network semantic communication. When driving at high speeds on highways, drivers typically focus their vision forward, leading to side blind spots. Assisting drivers in detecting movements in these blind spots, especially from swiftly approaching vehicles, is essential for safety. Vehicular network semantic communication technology can detect potentially hazardous vehicles in these blind spots. It achieves this by capturing, encoding, and transmitting real-time imagery of these vehicles, and then decoding and presenting this information as images to the driver. However, vehicular network semantic communication often grapples with bandwidth limitations, making effective image transmission a challenge.

To overcome this challenge, in this paper, we propose a scalable AIGC encoder-decoder architecture that extracts task-specific image semantics and transmits them in the form of text through the Internet of Vehicles \cite{9088328}. During this process, reinforcement learning techniques \cite{DBLP:conf/nips/Hasselt10, DBLP:conf/aaai/HasseltGS16} enhance the textual representation of the semantic information. If bandwidth permits, image regions with significant semantics are also conveyed. The ultimate objective is to refine and evaluate the quality of the reconstructed image until it meets acceptance criteria. The key contributions of this paper include:

\begin{enumerate}[leftmargin=*]
    \item We introduce a scalable AIGC encoder-decoder architecture that primarily transforms images into semantic textual information. Depending on bandwidth availability, it can additionally include relevant semantic image data. Our approach offers a twofold benefit: when bandwidth is constrained, it prioritizes transmitting semantic textual information, and when bandwidth is ample, it incorporates local image regions with significant semantics.
    \item We utilize reinforcement learning techniques to optimize encoding and decoding processes. By treating encoding and decoding as sequential decisions, we ensure the generated textual data retains ample semantic information, aiming to maximize the quality of the reconstructed image.
    \item We conduct experiments using a vehicle image dataset to validate the effectiveness of our proposed AIGC method. The results illustrate that our proposed method surpasses the baseline in both image quality and compression rate, suggesting its potential usefulness for real-world scenarios such as vehicular communication under varying bandwidth conditions.
\end{enumerate}

\begin{figure*}[t]
    \centering
    \includegraphics[width=0.67\textwidth]{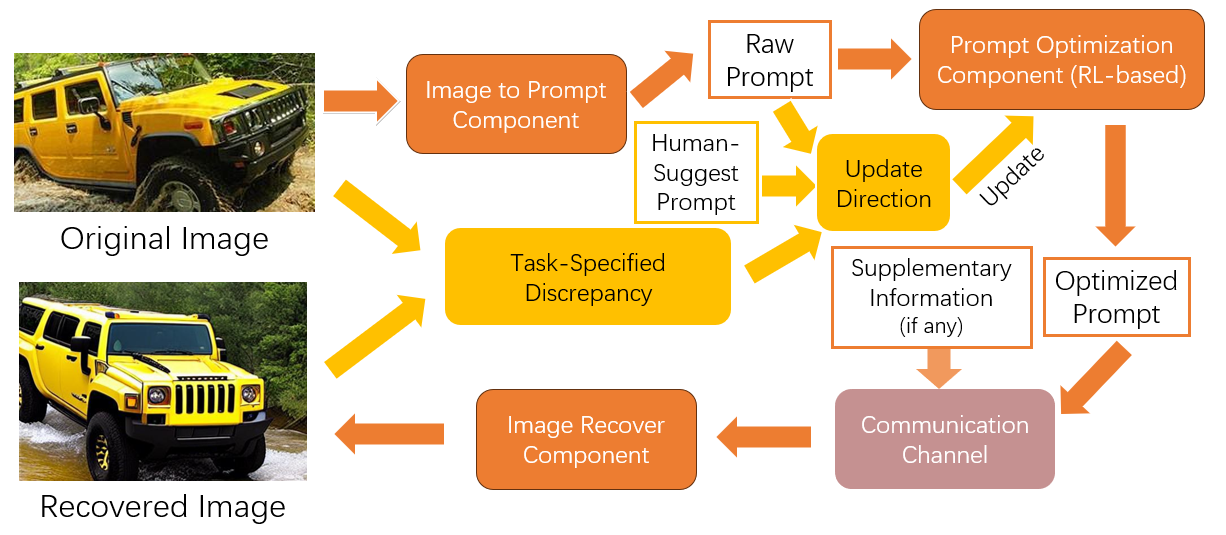}
    \vspace{-5pt}
    \caption{The architecture of the proposed scalable AIGC}
    \label{fig:arch_SCAI}
\end{figure*}

\section{Related Work}

\subsection{Image Compression and Transmission}

Conventional image compression methods like JPEG \cite{wallace1992jpeg} and PNG \cite{boutell1997png} are extensively used to decrease image size while maintaining acceptable quality. Recent advancements utilize deep learning architectures, like Recurrent Neural Networks (RNNs) \cite{DBLP:journals/corr/TodericiOHVMBCS15} and autoencoders \cite{DBLP:conf/iclr/TheisSCH17}, to attain superior compression rates without compromising image fidelity.

\subsection{Textual Descriptions from Visual Data}

Transforming images into textual descriptions or prompts has gained traction in research, particularly since the rise of deep learning. Initial efforts centered on template-driven techniques \cite{farhadi2010every}, while more recent approaches utilize RNNs \cite{vinyals2015show} and Transformers \cite{vaswani2017attention} to generate more natural and descriptive captions for images.

\subsection{Text-to-Image Synthesis}

The reverse challenge of transforming textual descriptions back into images has also attracted considerable interest. Generative Adversarial Networks (GANs) \cite{goodfellow2014generative} have been at the forefront of this research, with models like DALL·E \cite{ramesh2021zero} demonstrating the capability to generate high-quality images from textual prompts. Integrating supplementary cues or context to direct the synthesis process has been explored, bolstering the accuracy and relevance of the produced images \cite{zhang2017stackgan}.

\subsection{Reinforcement Learning in Image Processing}

The application of reinforcement learning in image processing tasks, such as optimization and enhancement, is a relatively new avenue. Works like \cite{mnih2015human} have shown the potential of RL-based methods in achieving superior results compared to traditional techniques, especially in scenarios where the objective is not explicitly defined.

\section{Methodology}

The proposed scalable AIGC system represents a paradigm shift in how we approach data transmission and semantic communication, especially in bandwidth-constrained environments. The scalable AIGC system dynamically adapts to bandwidth availability, prioritizing the transmission of essential information. This adaptability stems from sophisticated encoding and reinforcement learning optimization, enabling the scalable AIGC system to decide what and how to transmit.

The motivation of the scalable AIGC system is that transmitting high-resolution images isn't always practical or necessary. Often, a succinct textual representation capturing the image's essence is adequate. By transforming images into communication data, the scalable AIGC system ensures efficient transmission while retaining the data's semantic value.

The core of the proposed scalable AIGC system is to convert image data into a task-specific and more compact communication data format. Once transmitted, this data is optimized using reinforcement learning before being decoded into an image. The reinforcement learning method can force the communication data to be more related to the specific task. There are three main phases in our proposed method:

\begin{itemize}[leftmargin=*]
    \item \textbf{Information Encoding.} The initial phase involves encoding the image into a textual representation. This process, which we term as ``information encoding", leverages an encoder that distills the essential features of an image into a concise textual format suitable for transmission.
    \item \textbf{Reinforcement Learning-based Optimization.} Before decoding, the communication data is passed through a reinforcement learning module. This module identifies and adjusts the textual information, ensuring that the decoded image aligns well with the intended context and requirements. Specifically, using the actor-critic method, the model discerns detrimental textual details and recognizes phrases that can enhance the final image representation.
    \item \textbf{Information Decoding.} The optimized communication data is then fed into a decoder, translating the textual information back into its visual form, resulting in an image that is both bandwidth-efficient and contextually relevant.
\end{itemize}

As illustrated in Figure \ref{fig:arch_SCAI}, the scalable AIGC system encompasses three distinct components: the Image to Prompt Component, the Prompt Optimization Component, and the Image Recover Component. These components correspond respectively to the three stages previously described. In the subsequent sections, we will delve into a detailed discussion of these components respectively.

\vspace{-6pt}

\subsection{Information Encoding and Decoding}

\vspace{-2pt}

\subsubsection{Information Encoding}

The process of information encoding in our system leverages the capabilities of large language models. Given an image, the primary objective of this phase is to generate a concise and descriptive textual prompt that encapsulates the essential features and details of the image. This is achieved by feeding the image into our encoder, which has been trained on a vast dataset of image-text pairs. The underlying principle is to harness the power of state-of-the-art language models to distill the rich visual information of an image into a compact textual representation. This representation, termed as the ``prompt", serves as a bridge between the visual and textual domains, ensuring that the core semantics of the image are preserved in a bandwidth-efficient manner.

\vspace{-4pt}

\subsubsection{Information Decoding}

The decoding phase is tasked with the conversion of the generated textual prompt back into a visual representation. This is not a straightforward translation, as the challenge lies in regenerating an image that is both contextually relevant and closely resembles the original image. Our decoder employs advanced text-to-image synthesis techniques to achieve this.

Furthermore, to enhance the accuracy and fidelity of the regenerated image, our system allows the integration of image hints. These hints provide additional context and guidance to the decoder, ensuring that the output image aligns well with the original context. For instance, if the original image was of a sunset over a mountain range, the hint might emphasize the color palette or the silhouette of the mountains. By incorporating these hints, our decoder can produce images that are not only semantically aligned with the textual prompt but also visually congruent with the original image.

In essence, our encoding and decoding methodologies ensures a seamless transition between the visual and textual domains, paving the way for efficient and semantically rich communication in bandwidth-constrained scenarios.

\vspace{-4pt}

\subsection{Reinforcement Learning-based Optimization}

The process of compressing the rich details of an image into a textual representation, termed as ``information encoding", may not always capture the nuances vital for specific tasks and application scenarios. For instance, when encoding the details of surrounding vehicles, users are primarily concerned with specific aspects such as the direction from which the vehicle is approaching, the orientation of its front, and its type (be it a large truck, a sedan, or a small three-wheeled electric vehicle).

To bridge this gap between model-generated content and user-centric requirements, we propose a reinforcement learning-based approach to enhance the expressive capability of the encoded textual information. The objective is to seamlessly integrate details that are highly pertinent to driving contexts into the generated text, thereby elevating the model's performance.

The primary objective of employing the reinforcement learning method in our context is twofold:

\begin{enumerate}[leftmargin=*]
    \item \textbf{Identification of Detrimental Information.} Recognize and pinpoint textual elements within the encoded communication data that might be counterproductive or irrelevant to the overarching task.
    \item \textbf{Infusion of Beneficial Phrases.} Detect and suggest phrases or details that can significantly enhance the model's output in terms of contextual relevance and accuracy.
\end{enumerate}

By achieving these objectives, the model aims to eliminate detrimental textual details and incorporate beneficial information, ensuring that the decoded visual representation is both contextually rich and aligned with user preferences.

We use the actor-critic framework. 
The \emph{state} $s$ is defined as the current textual representation. The initial state is the communication data generated from the input image. 
The possible \emph{action}  $a$ is one of \emph{addition}, \emph{deletion} and \emph{modification}. The three kinds of actions can introduce a new phrase or detail into the communication data, remove a specific phrase or detail from the communication data, and alter an existing phrase or detail within the communication data, respectively. 
The \emph{reward} $r$ is a measure of the quality of the adjusted communication data in terms of its ability to be decoded into a contextually relevant image. 
Where the quality can be a function of various factors like contextual relevance, clarity, and alignment with user preferences.

There is an actor and a critic in the framework. 
The actor defines a policy \(\pi(a|s; \theta)\) which gives the probability of taking action \(a\) in state \(s\). It's represented as:

\vspace{-5pt}

\[ \pi(a_t|s_t; \theta) = P(a_t|s_t; \theta). \]

\begin{figure*}
    \centering
  \begin{minipage}[t]{0.3\linewidth}
    \centering
    \includegraphics[width=\linewidth]{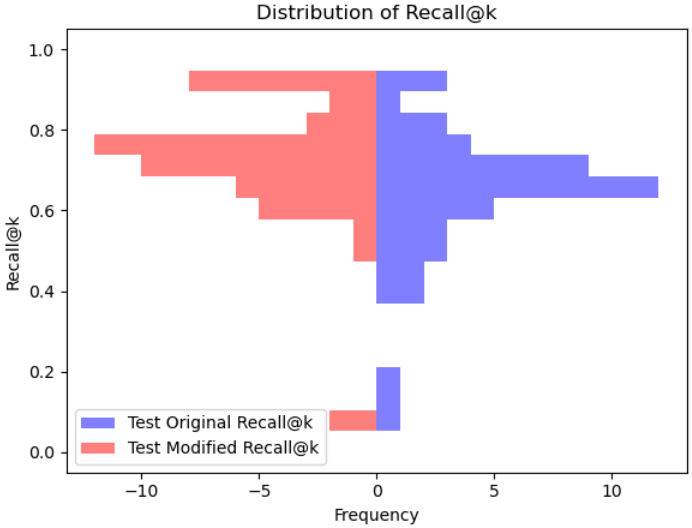}
    \caption{Distribution of Recall@k}
    \label{fig:distribution}
  \end{minipage}
  \hfill
  \begin{minipage}[t]{0.305\linewidth}
    \centering
    \includegraphics[width=\linewidth]{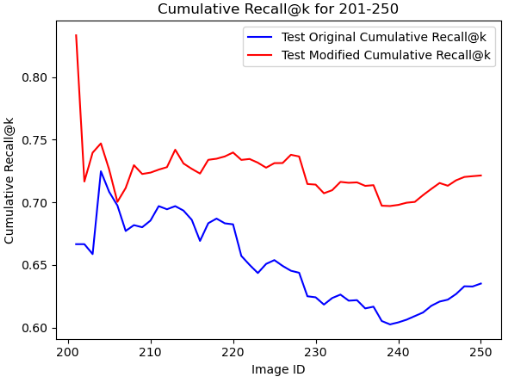}
    \caption{Cumulative Recall@k}
    \label{fig:cumu}
  \end{minipage}
  \hfill
  \begin{minipage}[t]{0.335\linewidth}
    \centering
    \includegraphics[width=\linewidth]{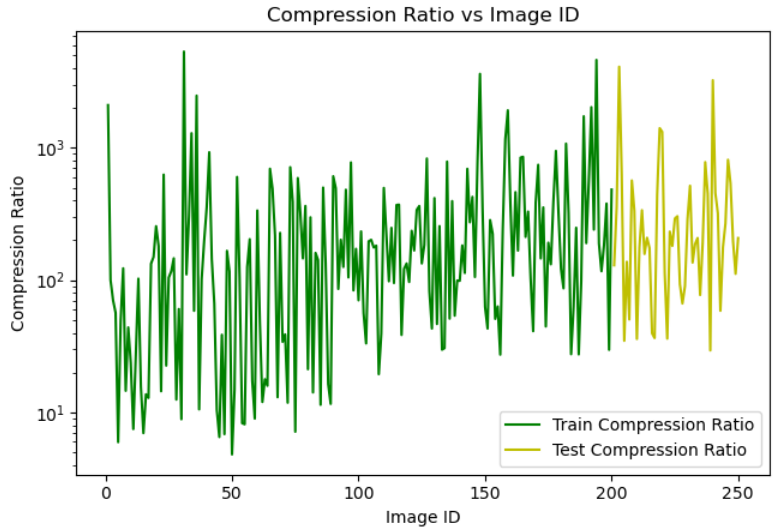}
    \caption{Compression Ratio}
    \label{fig:compression}
  \end{minipage}
\end{figure*}

The critic evaluates the expected return or value of taking an action in a particular state. It's given by:

\vspace{-5pt}

\[ V(s_t; \phi) = \mathbb{E}[R_t|s_t; \phi]. \]

The advantage function measures the relative value of taking a particular action over the average action in that state:

\vspace{-5pt}

\[ A(s_t, a_t) = r(s_t, a_t) + \gamma V(s_{t+1}; \phi) - V(s_t; \phi) \]

\noindent where \(\gamma\) is the discount factor.

The actor is updated using the policy gradient method:

\[ \nabla_\theta J(\theta) = \mathbb{E}[A(s_t, a_t) \nabla_\theta \log \pi(a_t|s_t; \theta)] . \]

The critic is updated based on the mean squared error between its predicted value and the actual return:

\vspace{-5pt}

\[ \nabla_\phi L(\phi) = \mathbb{E}[(r(s_t, a_t) + \gamma V(s_{t+1}; \phi) - V(s_t; \phi))^2]. \]

\section{Experiment}

In the experiment, we utilize the Stanford Cars dataset for our experiments. This dataset comprises 196 classes of cars, totaling 16,185 images. The categorization is typically based on the Make, Model, and Year of the cars. Each image has dimensions of 360×240.

We consider a method that immediately translates an image to text and then back to an image as our baseline. We compare the performance of the scalable AIGC system against this baseline. The experiments are conducted using i9-10920X CPU and GeForce RTX 2080 Ti. The evaluation is based on the Recall@k metric on the reconstructed images. This metric evaluates the accuracy of the top-k predictions for the classification task on the test set, making it suitable for assessing the performance of our image reconstruction.

We report results using a subset of the Stanford Cars dataset: 200 images for training and 50 images for testing. 

\begin{figure}
    \centering
    \includegraphics[width=0.45\textwidth]{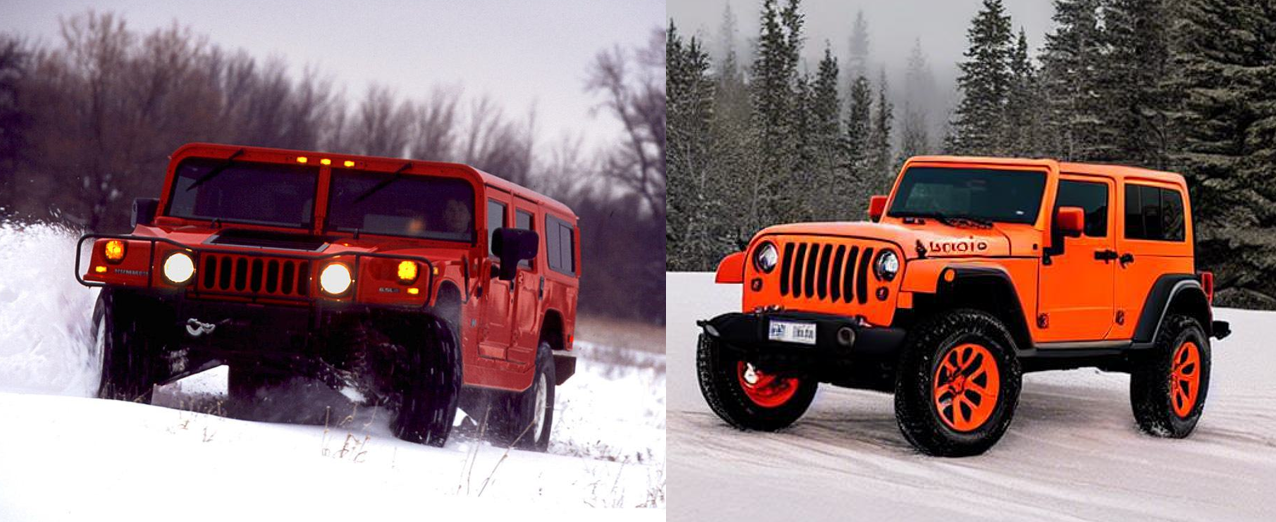}
    \vspace{-7pt}
    \caption{Example of Recovered Image of the Proposed Scalable AIGC System (Right) and the Original Image (Left).}
    \label{fig:generated}
\end{figure}

\textbf{Performance Distribution:} Figure \ref{fig:distribution} shows the distribution of the Recall@k metric for both our method (the scalable AIGC system, denoted as ``Modified'') and the baseline (denoted as ``Original'') on the same test set. Figure \ref{fig:cumu} shows the comparision between the cumulative Recall@k metrics of the proposed method and the baseline. From the presented figures, the proposed scalable AIGC system demonstrates some advantages over the baselines, suggesting good performance on specific tasks.

\textbf{Compression Rate:} Figure \ref{fig:compression} highlights the compression rate achieved by our method. Given that the proposed scalable AIGC system transmits adjusted textual information rather than the original image, it realizes substantial compression. This particularly advantageous in bandwidth-constrained situations, ensuring swift information transfer without compromising the semantic integrity of the reconstructed image. Therefore, the scalable AIGC system is a promising approach for bandwidth-constrained communication scenarios. Figure \ref{fig:generated} shows an example of the recovered image derived solely from text representations via the proposed  system and the corresponding original image. As evident from the figure, the images retain visual semantic consistency before and after processing.

\section{Conclusion}

In this paper, we present a scalable AI generative content system—a approach to efficient semantic communication in bandwidth-limited settings. Utilizing encoder-decoder architecture and reinforcement learning, our method encodes the image into a leaner way for transmission and reconstruction at the decoder side. Experimental tests on a vehicle image dataset confirm that our framework compresses raw images into task-specific textual representations. It then produces high-quality images from these textual cues, as evidenced by the cumulative Recall@k metric. These outcomes illustrate the efficacy and promise of our scalable system, which we believe holds versatility for application in various fields.

\vfill\pagebreak



\end{document}